%% file: main_arxiv.tex
\documentclass{article}

 \usepackage[main, final]{neurips_2025}
\usepackage[utf8]{inputenc} % allow utf-8 input
\usepackage[T1]{fontenc}    % use 8-bit T1 fonts
\usepackage{hyperref}       % hyperlinks
\usepackage{url}            % simple URL typesetting
\usepackage{booktabs}       % professional-quality tables
\usepackage{amsfonts}       % blackboard math symbols
\usepackage{nicefrac}       % compact symbols for 1/2, etc.
\usepackage{microtype}      % microtypography
\usepackage{xcolor}         % colors
\usepackage{xspace}
\usepackage{algorithm}
\usepackage{algpseudocode}
\usepackage{makecell}
\usepackage{multirow}
\usepackage{amsmath}
\usepackage{graphicx}
\usepackage{subcaption}
\input{preamble}

\title{\our: A Temporal Sparse Attention for Diffusion Transformers}

\newcommand{\our}{\ensuremath{\tt LiteAttention}\xspace}

\author{
  Dor Shmilovich \quad Tony Wu \quad Aviad Dahan \quad Yuval Domb \\
  MoonMath.ai \\
  {\tt\small research@moonmath.ai}\\
  \url{https://github.com/moonmath-ai/LiteAttention}
}

\begin{document}

\maketitle

\begin{abstract}
Diffusion Transformers, particularly for video generation, achieve remarkable quality but suffer from quadratic attention complexity, leading to prohibitive latency. 
Existing acceleration methods face a fundamental trade-off: dynamically estimating sparse attention patterns at each denoising step incurs high computational overhead and estimation errors, while static sparsity patterns remain fixed and often suboptimal throughout denoising. 
We identify a key structural property of diffusion attention, namely, its sparsity patterns exhibit strong temporal coherence across denoising steps. 
Tiles deemed non-essential at step $t$ typically remain so at step $t+\delta$. 
Leveraging this observation, we introduce \our, a method that exploits temporal coherence to enable evolutionary computation skips across the denoising sequence. 
By marking non-essential tiles early and propagating skip decisions forward, \our eliminates redundant attention computations without repeated profiling overheads, combining the adaptivity of dynamic methods with the efficiency of static ones. 
We implement a highly optimized \our kernel on top of FlashAttention and demonstrate substantial speedups on production video diffusion models, with no degradation in quality. 
\end{abstract}

\input{src/1-Introduction}
\input{src/2-RelatedWork}
\input{src/3-Preliminary}
\input{src/4-Method}
\begin{figure*}[h]
    \centering
    \includegraphics[width=.99\textwidth]{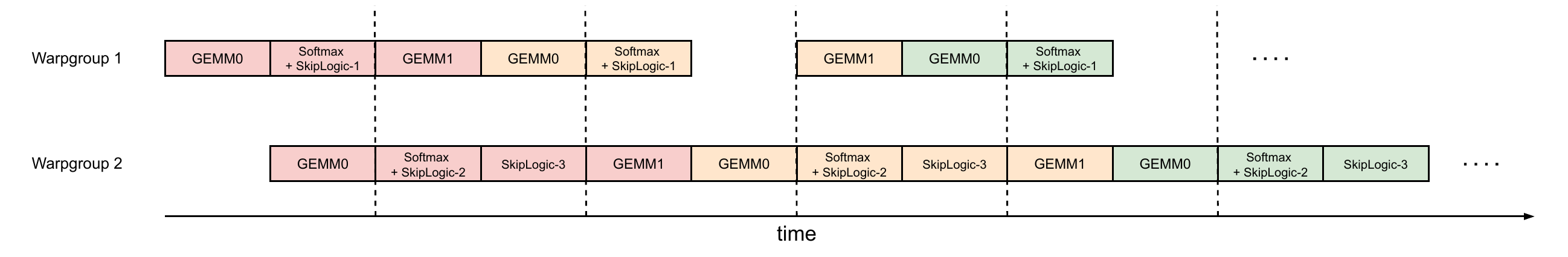} 
    % \vspace{-1em}
    \caption{\our's pipeline for the two warpgroup H100 configuration (based on FA3). In \textit{SkipLogic-1}, a skip bit is computed per each warp in the warpgroup. In \textit{SkipLogic-2}, a skip bit is again computed per warp and the result is combined with the bitmap of warpgroup 1. In \textit{SkipLogic-3} the warp-level skip bitmap is reduced to a single skip bit per the complete tile.}
    \label{fig:la_pipeline} 
\end{figure*}

\section{Implementation}

\subsection{\our}
\our is implemented atop FlashAttention3 (FA3)~\cite{dao2023flashattention}. 
Rather than re-implementing the kernel from scratch, we extend the FA3 API with an additional parameter, a configurable skip threshold that governs sparsity selection. 
\our maintains a persistent \textit{Skip-Mask} that records and reuses tile-level skip decisions across diffusion timesteps. 
These lightweight extensions preserve full compatibility with FA3 while introducing evolutionary sparsity, enabling skip patterns to propagate across successive transformer timesteps.

FA3 computes attention in fixed-size tiles along both the query and key/value dimensions. 
The tile geometry depends on the data type and the available shared-memory capacity. 
In our setup, \our targets the NVIDIA H100 (Hopper) GPU under CUDA~12.8, adopting the same configuration as the official FA3 Hopper implementation. 
Specifically, for \texttt{BF16}, each tile spans $128 \times 176$ elements per head for a head dimension of $128$, maximizing occupancy under Hopper’s $228$KB per-SM shared-memory limit. 
Each CUDA thread block comprises three warpgroups: one producer and two consumers operating in a pipelined fashion. 
The producer asynchronously streams $K_j$ and $V_j$ tiles from global to shared memory, while the consumer warpgroups process distinct query tiles $Q_i$ in parallel, performing the matrix multiplication $Q_i K_j^{\top}/\sqrt{d}$, executing the online softmax reduction for numerical stability, and multiplying by $V_j$ to accumulate partial outputs. 
This producer–consumer pipeline achieves near-complete overlap between memory transfers and computation. \our 's sparsity mechanism operates at the same tile granularity as FA3, allowing its skip logic to integrate seamlessly into the existing producer–consumer pipeline with minimal kernel modifications. 
The key additions occur within both the consumer and producer warpgroups: the consumer warpgroups evaluate the skip condition concurrently with the online softmax computation, while the producer warpgroup consults the \textit{Skip-List} to stream only the relevant $K_j$ and $V_j$ tiles.

\subsection{Skip-Mask Evaluation Mechanism}
\begin{figure}[!htbp]
    \centering
    \begin{subfigure}[t]{0.49\columnwidth}
        \centering
        \includegraphics[width=\columnwidth]{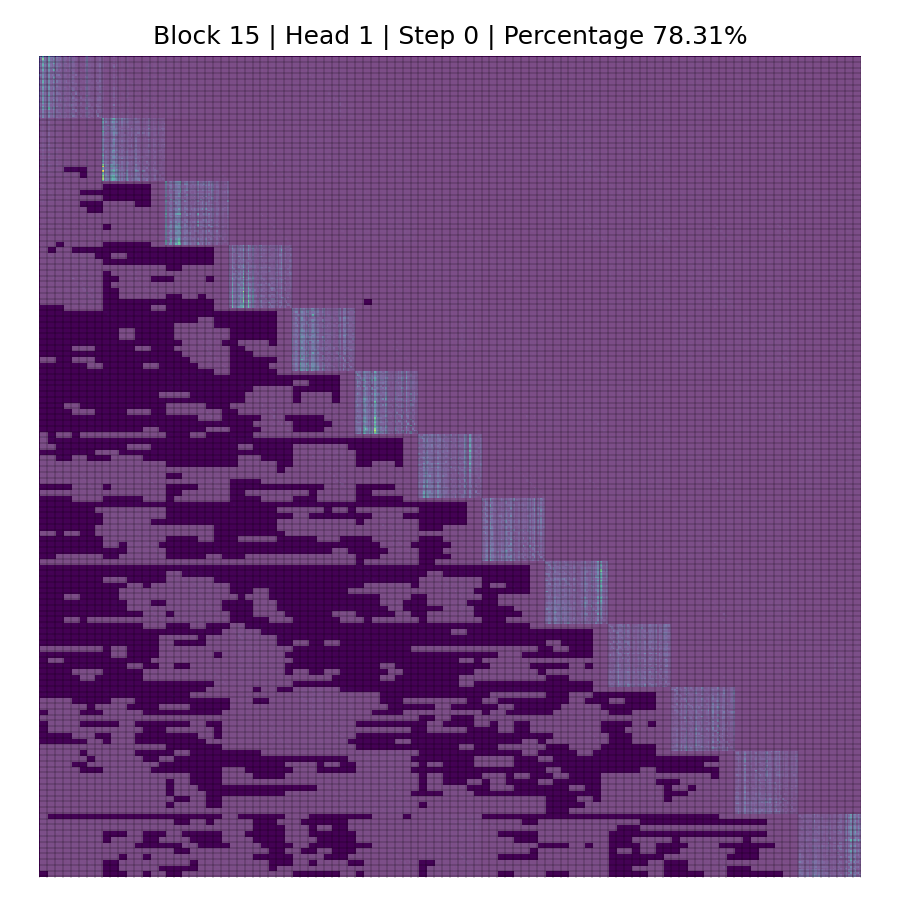} 
    \end{subfigure}
    \begin{subfigure}[t]{0.49\columnwidth}
        \centering
        \includegraphics[width=\columnwidth]{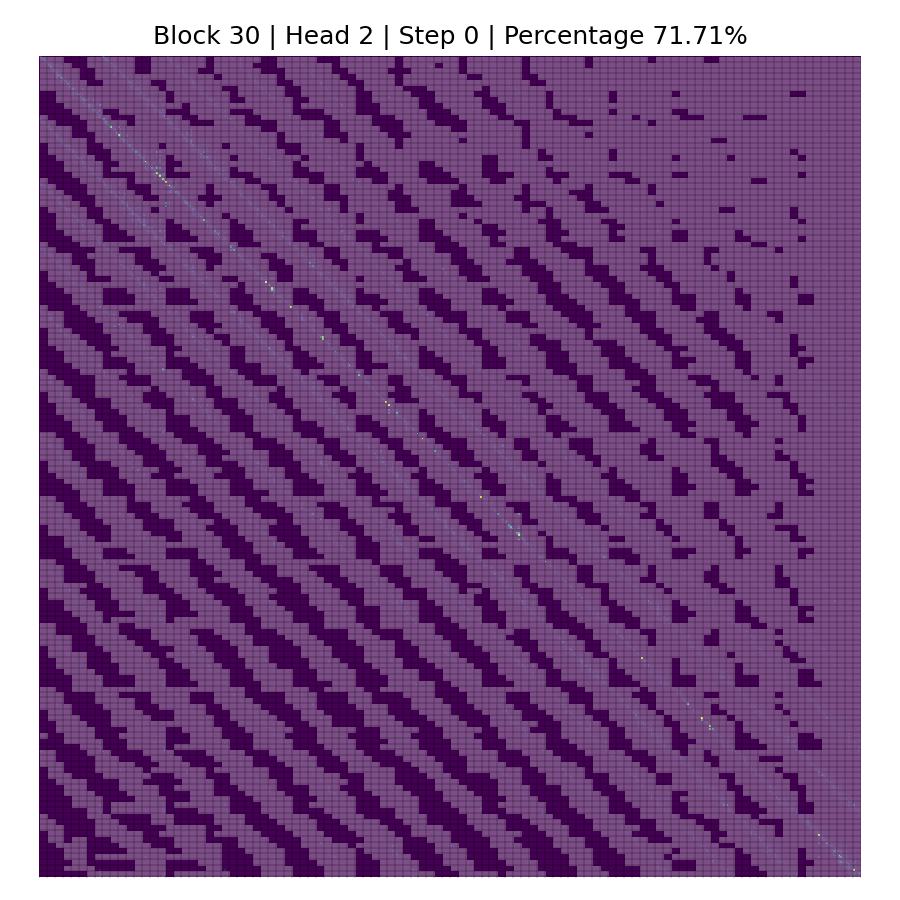} 
    \end{subfigure}
    \begin{subfigure}[t]{0.49\columnwidth}
        \centering
        \includegraphics[width=\columnwidth]{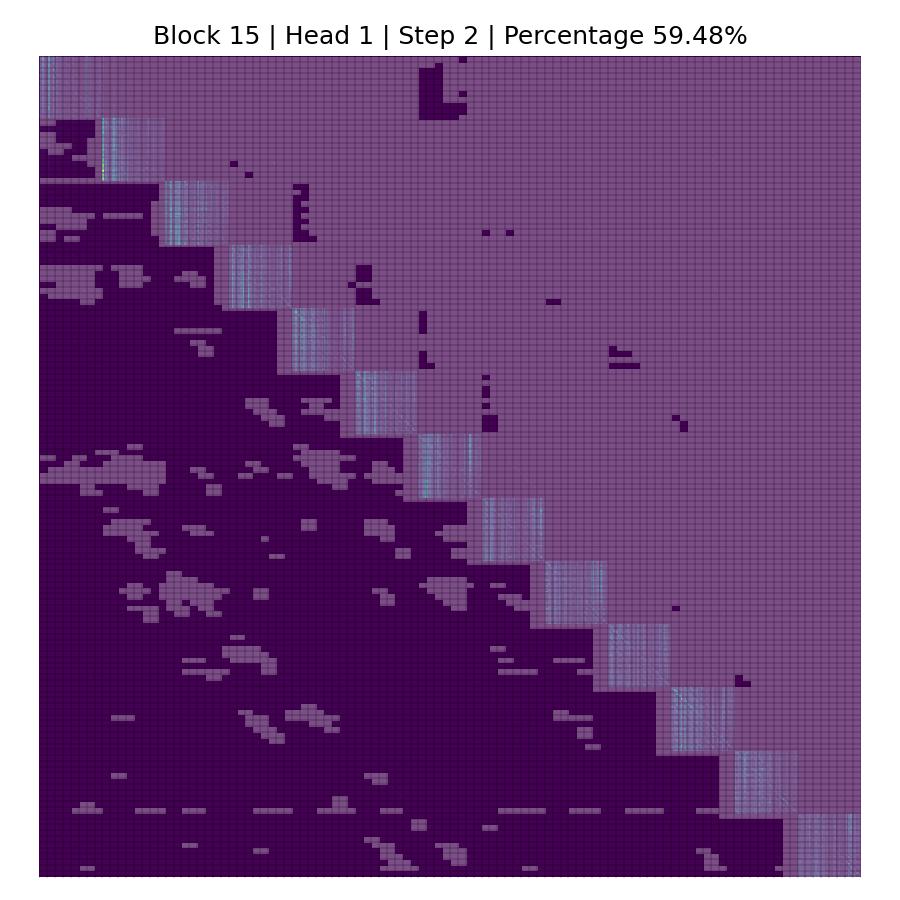} 
    \end{subfigure}
    \begin{subfigure}[t]{0.49\columnwidth}
        \centering
        \includegraphics[width=\columnwidth]{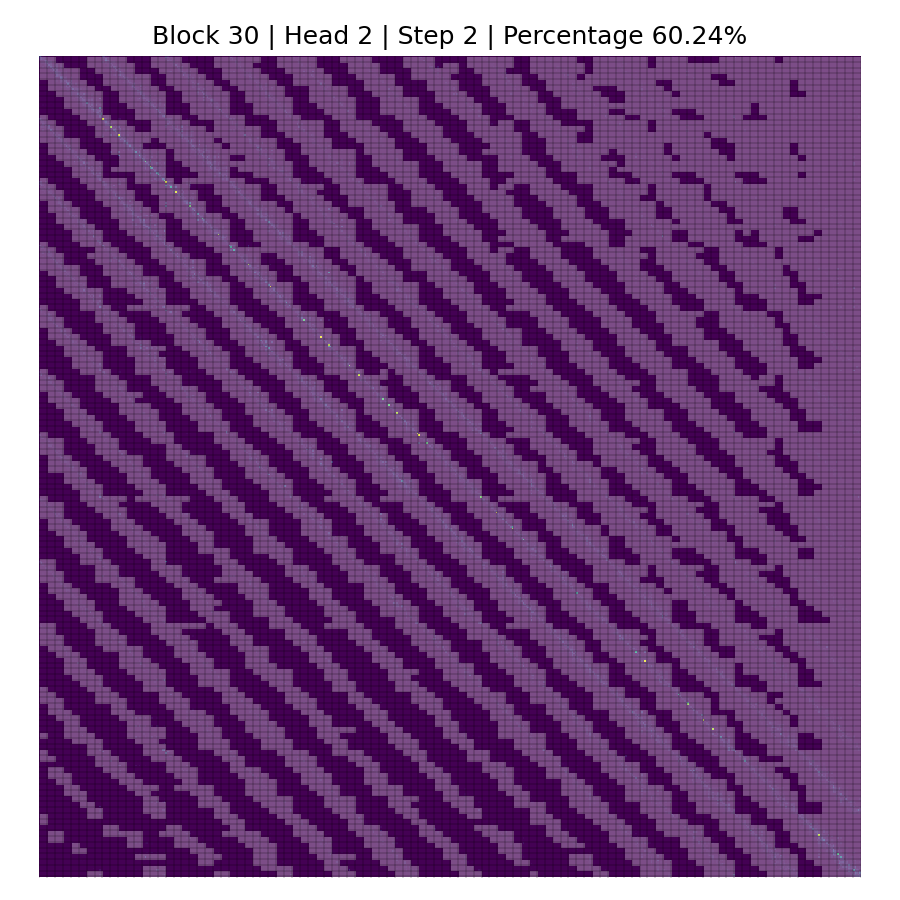} 
    \end{subfigure}
    \begin{subfigure}[t]{0.49\columnwidth}
        \centering
        \includegraphics[width=\columnwidth]{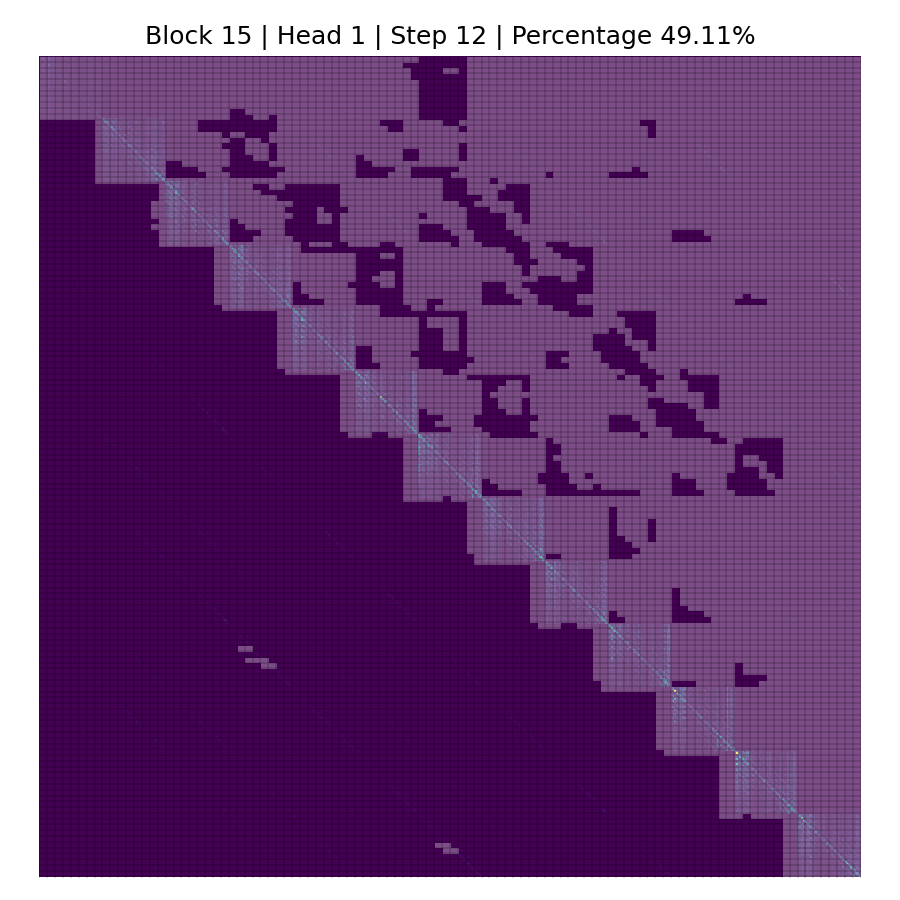} 
        \caption{Block 15 - Head 1}
    \end{subfigure}
    \begin{subfigure}[t]{0.49\columnwidth}
        \centering
        \includegraphics[width=\columnwidth]{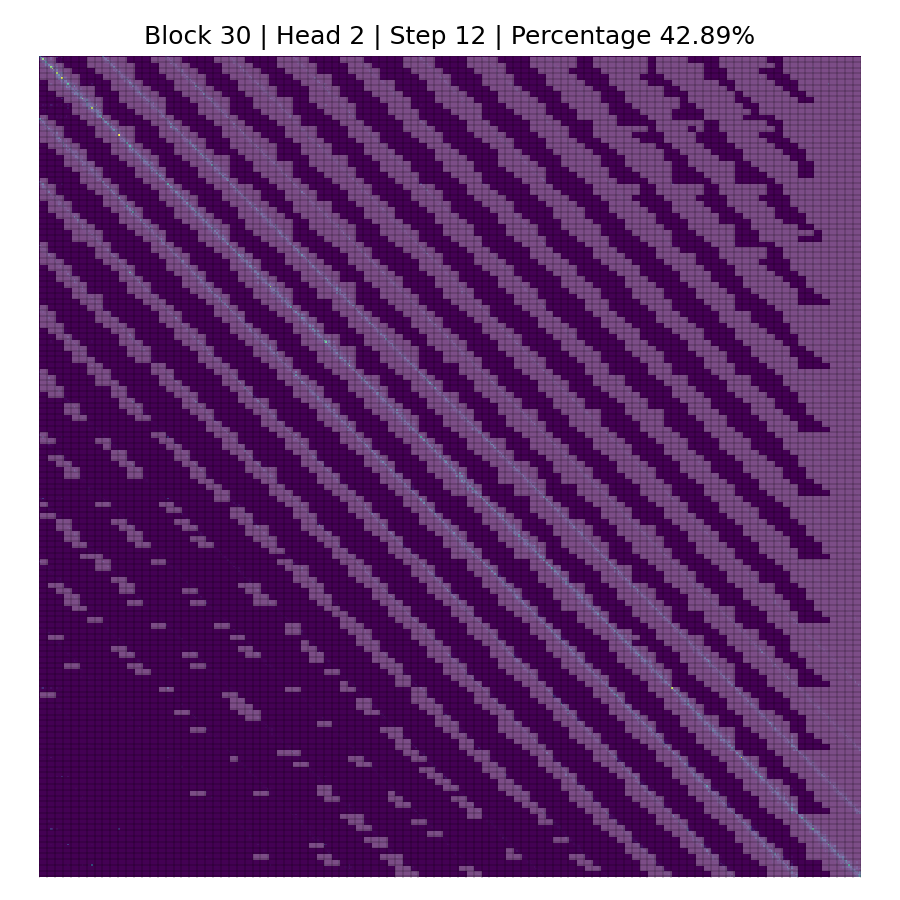} 
        \caption{Block 30 - Head 2}
    \end{subfigure}
    \caption{The evolving \textit{Skip-Mask} across diffusion timesteps for LTX-13B \cite{HaCohen2024LTXVideo} over two block/head sets. The top and bottom are the start and end masks, respectively. Dark purple means skipped.}
    \label{fig:skip_mask} 
\end{figure}

For each $QK$ tile, \our evaluates the skip condition \eqref{eq:sparge_cond}, which determines whether the tile’s contribution to the output is negligible. 
Each per-tile decision is written to a global \textit{Skip-Mask} data structure that persists across denoising timesteps, as depicted in Figure \ref{fig:skip_mask}.
Although this predicate could, in principle, be used to immediately bypass the subsequent $PV$ computation, doing so would introduce additional latency on Hopper due to synchronization dependencies within the consumer warpgroups. 
On the H100, each matrix multiply–accumulate (MMA) operation is executed by a warpgroup of four warps, and enforcing cross-warp agreement on the skip predicate before continuation would stall otherwise overlapping execution. 
To avoid this, \our records partial skip results independently for each of the four warps. 
Intra-warp reductions are implemented efficiently using warp-synchronization primitives, while cross-warp reductions, which require barriers, are deferred to the kernel epilogue, executed once by the final warpgroup after completing all softmax operations for the current tile.
Figure \ref{fig:la_pipeline} depicts \our's pipeline, constructed on top of FA3 (see \cite{shah2024flashattention} - Figure 1).

\begin{table*}[h!]
\centering
\caption{Comparison of \our's video quality (VBench), attention sparsity (Sps), and runtime (Run) compared with FlashAttention3 (FA3), SparseVideoGen, and RadialAttention over Wan2.1-14B and Wan2.2-14B.
Best results are in \textbf{bold} and second best in \textit{italic}.}
\begin{tabular}{lccccccccccc}
\toprule
\textbf{Method} & \textbf{AQ}$\uparrow$ & \textbf{BC}$\uparrow$ & \textbf{DD}$\uparrow$ & \textbf{IQ}$\uparrow$ & \textbf{SC}$\uparrow$ & \textbf{TF}$\uparrow$ & \textbf{TS}$\uparrow$ & \textbf{Sps[\%]}$\uparrow$ & \textbf{Run[sec]}$\downarrow$ \\
\toprule
\textbf{Wan2.1-14B} \\
\midrule
FA3 & 0.676 & 0.977 & 0.417 & 68.74 & 0.965 & 0.962 & 0.137 & 0 & 1707 \\
\midrule
SVG & \textit{0.665} & \textit{0.971} & \textbf{0.500} & \textbf{68.58} & 0.962 & 0.959 & \textit{0.066} & \textit{66} & \textit{1019} \\
Radial & 0.660 & 0.970 & \textit{0.417} & 64.73 & \textbf{0.964} & \textbf{0.972} & 0.061 & \textbf{74} & 1192\\
\ensuremath{\tt{Lite}} & \textbf{0.677} & \textbf{0.975} & \textbf{0.500} & \textit{66.76} & \textit{0.963} & \textit{0.962} & \textbf{0.142} & 42 & \textbf{902} \\
\toprule
\textbf{Wan2.2-14B} \\
\midrule
FA3 & 0.693 & 0.977 & 0.583 & 72.73 & 0.970 & 0.953 & 0.133 & 0 & 1473 \\
\midrule
SVG & \textit{0.689} & 0.962 & \textit{0.417} & \textit{72.24} & 0.961 & \textit{0.952} & \textit{0.061} & \textbf{66} & \textit{1022}\\
Radial & 0.682 & \textit{0.974} & \textbf{0.500} & \textbf{72.73} & \textit{0.967} & 0.947 & \textit{0.061} & \textbf{66} & 1207\\
\ensuremath{\tt{Lite}} & \textbf{0.698} & \textbf{0.977} & \textbf{0.500} & 71.44 & \textbf{0.969} & \textbf{0.953} & \textbf{0.135} & \textit{32} & \textbf{893} \\
\bottomrule
\end{tabular}
\label{tab:compare_results}
\end{table*}

\begin{table*}[h!]
\centering
\caption{Ablation study of \our's self-attention runtime (SR), its relative improvement (dSR), and video quality (VBench) over different levels of induced sparsity (Sps).}
\begin{tabular}{cccccccccccc}
\toprule
\textbf{Sps[\%]} & \textbf{SR[sec]}$\uparrow$ & \textbf{dSR[\%]}$\uparrow$ & \textbf{AQ}$\uparrow$ & \textbf{BC}$\uparrow$ & \textbf{DD}$\uparrow$ & \textbf{IQ}$\uparrow$ & \textbf{SC}$\uparrow$ & \textbf{TF}$\uparrow$ & \textbf{TS}$\uparrow$ \\
\midrule
0 & 695 & 0 & 0.702 & 0.978 & 0 & 77.100 & 0.994 & 0.978 & 0.092 \\
21 & 573 & 18 & 0.692 & 0.977 & 0 & 76.860 & 0.993 & 0.978 & 0.098 \\
42 & 418 & 40 & 0.690 & 0.964 & 0 & 76.086 & 0.987 & 0.978 & 0.096 \\
57 & 308 & 56 & 0.672 & 0.962 & 0 & 76.393 & 0.969 & 0.976 & 0.094 \\
77 & 163 & 77 & 0.630 & 0.964 & 0 & 77.061 & 0.962 & 0.978 & 0.075 \\
\bottomrule
\end{tabular}
\label{tab:ablation_results}
\end{table*}

\subsection{Producer and Skip-List Optimization}
The resulting per-tile skip flags are consumed by the producer warpgroup in the subsequent diffusion timestep. 
Because the producer is lightweight and inherently synchronized to wait for the consumer’s completion, it can query the skip decision without impacting throughput. 
If a tile is marked as skippable, the producer omits loading the corresponding $K_j$ and $V_j$ tiles from global memory, allowing the consumer to bypass all computation for the entire $j$'th iteration.
Initially, the skip mask was implemented as a simple bitmask, allocating one bit per $S_{ij}$ tile. 
However, as sparsity increased, we observed that a compressed representation offered superior efficiency. 
We therefore adopted a \textit{Skip-List} structure inspired by run-length encoding, where consecutive non-skipped ranges are represented as \texttt{(start, end)} pairs. 
This compact format enables the producer to skip entire contiguous sequences of $K$ tiles with a single conditional check, improving both memory efficiency and kernel throughput under high sparsity conditions.

\input{src/6-Experiments}
\input{src/7-Conclusions}

% {
%     \small
%     \bibliographystyle{ieeenat_fullname}
%     \bibliography{main}
% }

{
    \small
    \bibliographystyle{ieeenat_fullname}

\input{main.bbl}
}

\end{document}

%% file: preamble.tex
\usepackage{algorithm}
\usepackage{algpseudocode}
\usepackage{makecell}
\usepackage{multirow}

%% file: src/1-Introduction.tex
\section{Introduction}
Video generation via Diffusion Transformers (DiT) has reached a remarkable inflection point: models now produce compelling, high-fidelity content that rivals professional production quality. 
Yet this capability conceals a fundamental inefficiency. 
Despite their impressive generative performance, these models demand extraordinary computational resources, generating a single 5-second video can take up to 30 minutes even on state-of-the-art GPUs. 
The primary culprit is the computationally quadratic attention \cite{vaswani2017attention} mechanism: in some video diffusion architectures, the attention mechanism alone accounts for up to 80\% of the total inference latency~\cite{sparse_vdit2025}. 

The computational burden of diffusion models has motivated extensive research into efficient inference strategies~\cite{ma2024deepcache_cvpr,liu2024faster,ma2024learning_neurips,tang2024adadiff_eccv}. 
Two main directions have emerged: 
(1) \textbf{Dynamic methods}, which exploit sparsity \emph{within} each denoising step~\cite{xi2025svg,li2025radial,zhang2025spargeattn,zhang2025sliding}; and 
(2) \textbf{Static and caching methods}, which exploit redundancy \emph{across} denoising steps~\cite{zou2025token_iclr,bu2025dicacheletdiffusionmodel,lv2025fastercachetrainingfreevideodiffusion,liu2025timestepembeddingtellsits,chu2025omnicachetrajectoryorientedglobalperspective}. 
Dynamic methods repeatedly determine which computations to skip, incurring overhead and estimation noise, while static methods may misalign with evolving attention patterns. 
Critically, none exploit the temporal persistence of sparsity patterns across denoising steps. Our central observation is that tiles deemed non-essential at denoising step $t$ tend to remain non-essential at step $t+\delta$. 
This \textit{temporal coherence of sparsity} enables a fundamentally different strategy: identify skippable tiles once during early denoising, and propagate these skip decisions forward through the entire trajectory.

We present \our, which leverages temporal sparsity coherence to propagate computation skips through the denoising process.
By determining skip patterns early and reusing them throughout the denoising processs, \our achieves three key advantages simultaneously: 
(1) the \emph{content adaptivity} of dynamic sparsity (patterns are derived from actual attention statistics), 
(2) the \emph{efficiency} of static sparsity (no per-step re-evaluation overhead), and 
(3) the \emph{completeness} of full computation elimination. 
Together, these properties yield substantial acceleration while preserving the generative fidelity of DiT.

\subsection{Evolutionary Computation Skips}
\our performs \textit{evolutionary skips}, completely eliminating attention computation for tiles marked as skippable. 
Once a tile is skipped, the entire attention iteration for that tile is bypassed across subsequent timesteps where the skip decision remains valid. 
This full-iteration elimination distinguishes \our from prior sparse attention approaches that skip only partial attention computations, where major bottlenecks such as softmax evaluation and memory transfers continue to dominate runtime despite partial sparsification.

\subsection{Implementation and Robustness}
\our integrates seamlessly into modern CUDA-accelerated attention kernels via FlashAttention3, maintaining full compatibility while introducing only moderate memory overhead for storing metadata.
The method is production-ready and does not require model retraining or architectural modifications.

To ensure robustness when skip decisions persist across timesteps with varying denoising conditions, \our incorporates a lightweight calibration mechanism that weights approximation errors by their layer-dependent impact. 
This calibration acts as a supporting component to maintain accuracy, while the core contribution lies in the evolutionary skip mechanism.

% \subsection{Contributions}
\smallskip
\noindent Our key contributions are summarized as follows:
\begin{itemize}
    \item \textbf{Evolutionary skip framework.} We introduce a mechanism that exploits the temporal coherence of sparsity patterns to eliminate full attention computations for marked tiles across denoising timesteps.
    \item \textbf{Amortized sparsity profiling.} We determine which tiles can be skipped early in the denoising process and reuse these skip decisions for all subsequent timesteps, eliminating the need for repeated sparsity profiling.
    \item \textbf{Efficient GPU implementation.} We develop optimized CUDA kernels leveraging evolved skip masks with moderate memory overhead, achieving efficient runtime performance.
\end{itemize}

%% file: src/2-RelatedWork.tex
\section{Related Work}
\label{sec:related}
Accelerating diffusion models requires addressing their substantial computational demands. 
Prior work has approached this challenge along two orthogonal directions: reducing attention within individual denoising steps, or exploiting redundancy across the denoising sequence.

\subsection{Per Timestep Sparse Attention}
Recent work has observed that attention patterns exhibit significant sparsity within individual denoising steps. Methods in this category determine sparse patterns either statically or dynamically, but crucially, these determinations are made independently at each step.

Sparse VideoGen (SVG)~\cite{xi2025svg,yang2025sparse} presented dynamic sparse attention for video DiTs by classifying attention heads into spatial and temporal categories and profiling sparse patterns within each step. Sparse-vDiT~\cite{sparse_vdit2025} complements this with architectural insights, showing that attention patterns follow recurring structures: diagonal blocks for self-frame interactions, multi-diagonal blocks for cross-frame consistency, and vertical stripes for global tokens. These patterns are largely input-invariant and intrinsic to the model architecture, yet their stability across timesteps is not exploited.

Radial Attention~\cite{li2025radial} formalizes Spatiotemporal Energy Decay, proposing static $O(n \log n)$ masks. While theoretically efficient, static patterns sacrifice adaptivity. SpargeAttention~\cite{zhang2025spargeattn} presents a sparse attention framework that predicts low-attention blocks via two-stage online filtering which requires profiling at every step. Sliding Tile Attention (STA)~\cite{zhang2025sliding} takes advantage of the concentration of attention at the tile-level through sliding windows in granularity of the tile.

% Sparse VideoGen (SVG)~\cite{xi2025svg,yang2025sparse} presented dynamic sparse attention for video DiTs by classifying attention heads and profiling sparse patterns within each step. Sparse-vDiT~\cite{sparse_vdit2025} showed that attention patterns follow recurring structures: diagonal blocks for self-frame interactions, multi-diagonal blocks for cross-frame consistency, and vertical stripes for global tokens, yet their stability across timesteps is not exploited. Radial Attention~\cite{li2025radial} proposes static $O(n \log n)$ masks with exponentially decaying compute density, while SpargeAttention~\cite{zhang2025spargeattn} predicts low-attention blocks via two-stage online filtering at every step. Sliding Tile Attention (STA)~\cite{zhang2025sliding} exploits tile-level concentration through sliding windows at tile granularity.

In every one of these approaches, sparsity patterns are determined independently at each denoising step, either via dynamic recomputation or static commitment, without leveraging their persistence across steps.

\subsection{Cross-Sequence Redundancy Exploitation}
Another line of research leverages structure across denoising timesteps, observing that different phases exhibit varying computational requirements.

TGATE~\cite{liu2024faster} noted that cross-attention outputs converge in early denoising steps, enabling phase-based computation where patterns are cached and reused later. DeepCache~\cite{ma2024deepcache_cvpr} demonstrates that high-level transformer features remain similar across adjacent timesteps, allowing feature caching without retraining. Learning-to-Cache (L2C)~\cite{ma2024learning_neurips} learns timestep-dependent layer-level routing strategies, identifying which layers benefit from caching throughout the denoising sequence. Token-wise Feature Caching~\cite{zou2025token_iclr} captures fine-grained temporal redundancy at the token level using layer-specific caching ratios. AdaDiff~\cite{tang2024adadiff_eccv} implements dynamic early exit based on timestep-aware uncertainty, allocating the computation proportional to the importance of each phase. While these approaches exploit cross-sequence structure in features or layer utilization, they do not consider temporal structure in attention sparsity patterns. Additionally, they retain or approximate intermediate values, incurring nontrivial estimation errors and significant memory overheads.

\subsection{Sparsity Stability Across Denoising: A New Paradigm}
Unlike per-step sparse attention methods, which recompute patterns at each denoising step, or cross-sequence methods that exploit feature redundancy, \our is based on a fundamentally different principle: attention sparsity patterns remain stable throughout the denoising sequence.

The key observation is that attention sparsity is temporally coherent rather than random. By exploiting this persistence, \our determines skip patterns once and reuses them across the entire trajectory, achieving full elimination of attention computation for marked tiles without repeated profiling. \our thus enables a new class of accelerators that combine adaptive per-timestep sparsity with zero repeated profiling. Concurrent work, SparseD~\cite{wang2025sparsed}, observes similar temporal sparsity stability in diffusion language models, applying cross-step sparsity in a different domain.

%% file: src/3-Preliminary.tex
\section{Preliminaries}

\subsection{Attention Mechanisms}
\label{ss:perm_attn}

\smallskip \noindent{\textbf{FlashAttention.}
FlashAttention~\cite{dao2023flashattention} is an efficient attention algorithm that significantly reduces the memory bandwidth requirements of standard attention implementations by employing a tiling-based memory strategy. 
It is specifically optimized for NVIDIA GPUs, though its principles naturally extend to other parallel-processing architectures.

Given the standard attention formulation
\begin{align}
    S &= QK^\top / \sqrt{d},\\
    P &= \sigma(S),\\
    O &= PV,
\end{align}
where $Q \in \mathbb{R}^{n \times d_q}$ and $K, V \in \mathbb{R}^{n \times d_k}$ are the query, key, and value matrices, $n$ denotes the sequence length, and $d_q, d_k$ are the hidden dimensions (for simplicity we set $d := d_q = d_k$).
The softmax operator $\sigma(\cdot)$ is applied row-wise such that $p_{ij} := \exp(s_{ij}) / \sum_k \exp(s_{ik})$ where $p_{ij}$ and $s_{ij}$ are elements of $P$ and $S$, respectively.

In FlashAttention, the matrices $Q$, $K$, and $V$ are partitioned along the sequence dimension into tile sets 
$\{Q_i\},~\{K_j\},~\{V_j\}$ where the tiles are of sizes $h_q \times w$, $h_k \times w$, and $h_k \times w$, respectively.
These tiles are processed sequentially using the \emph{online softmax} algorithm~\cite{milakov2018online}, which maintains numerical stability and enables incremental accumulation of partial results.
For each query tile $Q_i$, the computation proceeds iteratively over the key-value tile pairs $\{(K_j, V_j)\}$:
\begin{align}
    S_{ij} &= Q_i K_j^\top / \sqrt{d},\\
    (\widetilde{P}_{ij}, \mathbf{m}_{ij}) &= \tilde{\sigma}(S_{ij}, \mathbf{m}_{i,j-1}), \\
    \widetilde{O}_{ij} &= \widetilde{P}_{ij}V_j,
\end{align}
where $\widetilde{O}_{ij}$ is a partial output, $\widetilde{P}_{ij} := \exp(S_{ij} - \mathbf{m}_{ij})$, and $\mathbf{m}_{ij} := \mathrm{rowmax}(\mathbf{m}_{i,j-1}, \mathrm{rowmax}(S_{ij}))$, where $\mathrm{rowmax}(\cdot)$ operates on a matrix and outputs a vector of the maximums of all rows.
The value $\mathbf{m}_{ij}$ is updated cumulatively across the tile index $j$.
For brevity, the remainder of the online softmax procedure is omitted.

\smallskip \noindent{\textbf{SpargeAttention.}
The \emph{sparse online softmax}, introduced in SpargeAttention~\cite{zhang2025spargeattn}, extends FlashAttention with a dynamic pruning mechanism that skips partial computation of tiles whose contribution to the output is negligible.
Specifically, when the local maximum
$\mathbf{m}_{\text{local}} := \mathrm{rowmax}(S_{ij})$
is significantly smaller than the cumulative maximum $\mathbf{m}_{ij}$, the corresponding tile has exponentially suppressed weights.
The computation of tile $(K_j, V_j)$ is partially skipped when
\begin{align}
    \max(\mathbf{m}_{\text{local}} - \mathbf{m}_{ij}) \le -\varepsilon,
    \label{eq:sparge_cond}
\end{align}
for a chosen threshold $\varepsilon > 0$ (note that we are required to take a maximum since $\mathbf{m}$ is a vector of row maximums).
In this case, $\max(\widetilde{P}_{ij}) \le e^{-\varepsilon}$, implying that the term $\widetilde{P}_{ij}V_j$ contributes negligibly to the final output and can be safely omitted.
With an appropriately chosen threshold, SpargeAttention preserves the guarantees of the online softmax while eliminating redundant computation and improving efficiency.

\subsection{Flow, Diffusion Transformers, and Caching}

\smallskip \noindent{\textbf{Flow.}
We begin by defining an Ordinary Differential Equation (ODE) as
\begin{align}
    dX_t &= u(X_t, t)\,dt, \\
    X_0 &= x_0,
\end{align}
where $t \in [0,1]$, $u(x, t)$ is a vector field, and $x_0$ is an initial condition.
A solution to this ODE is termed a \emph{trajectory}, and the collection of all trajectories arising from all possible initial conditions constitutes a \emph{flow}.
When $X_t$ is regarded as a random process, the flow induces a mapping between two distributions, $p(X_0)$ and $p(X_1)$.
An alternative formulation, the Stochastic Differential Equation (SDE),\footnote{An SDE takes the form $dX_t = u(X_t, t)\,dt + \sigma_t\,dW_t$ where $W_t$ is Brownian motion.} gives rise to a \emph{diffusion} process instead of a deterministic flow.
In this work, we use the terms \emph{flow} and \emph{diffusion} interchangeably.

Consider $X_0 \sim p_{\mathrm{init}}(x)$ and $X_1 \sim p_{\mathrm{data}}(x)$, where $p_{\mathrm{init}}(x)$ is a simple noise distribution independent of the data distribution $p_{\mathrm{data}}(x)$.
A trajectory under this formulation maps a noise sample into a meaningful data sample.
A \emph{flow model} (also known as a \emph{flow matching})~\cite{lai2025principlesdiffusionmodels} parameterizes the vector field $u_\theta(x,t)$ with learnable parameters $\theta$, and is trained to reverse the forward process
\begin{align}
    X_t = \sqrt{\alpha_t} X_1 + \sqrt{1-\alpha_t} X_0,
\end{align}
where $0 \leq \alpha_t \leq 1$ is the \emph{noise schedule}.
In this work, we consider an extended form of the vector field,
\begin{align}
    u_\theta(x,t,c),
\end{align}
where $c$ denotes a conditioning signal that guides trajectories toward the conditional data distribution $p_{\mathrm{data}}(x \mid c)$.

\smallskip \noindent{\textbf{Diffusion Transformers.}
A DiT~\cite{peebles2023scalable} is a feedforward architecture composed of $M$ bidirectional transformer blocks (or layers).
Each block typically consists of a sequence of submodules: self-attention, cross-attention, and a multilayer perceptron (MLP).
See \cite{peebles2023scalable} and \cite{chen2024gentron} for common variants of transformer block architectures.
Let $\mathcal{T}_i$ denote the $i$-th transformer block in a DiT; its output is defined as the composition
\begin{align}
    y_t^i = \mathcal{T}_i \circ \mathcal{T}_{i-1} \circ \cdots \circ \mathcal{T}_1(x_t, t, c),
\end{align}
where $(t,c)$ serves as conditioning input to all transformer blocks and the final DiT output is $y_t := y_t^M$.
Our learned diffusion vector field $u_\theta(x,t,c)$ is implemented using this DiT backbone.

% \begin{figure}[h]
%     \centering
%     \includegraphics[width=.86\columnwidth]{src/figs/dit.pdf} 
%     % \vspace{-1em}
%     \caption{Transformer Block Architectures. \textit{Left:} Transformer with Cross-Attention. \textit{Right:} Transformer with In-Context Conditioning (a.k.a. Multimodal).}
%     \label{fig:dit} 
% \end{figure}

\smallskip \noindent{\textbf{Caching.}
Caching exploits the slow temporal evolution of DiT outputs across diffusion timesteps to improve computational efficiency.
Following~\cite{bu2025dicacheletdiffusionmodel}, define the residual at layer $i$ and timestep $t$ as
\begin{align}
    r_t^i := y_t^i - x_t.
\end{align}
In particular, the final residual $r_t := r_t^M$ often changes slowly between adjacent timesteps, i.e. $\|r_t - r_{t-1}\|$ is small.
This property can be exploited by caching and reusing $r_{t-1}$ instead of recomputing the full DiT output at each step.

Furthermore, it has been observed in~\cite{bu2025dicacheletdiffusionmodel} that a small relative change between successive intermediate representations,
\begin{align}
    \gamma_t^i:=\frac{\|y_t^i - y_{t-1}^i\|}{\|y_t^i\|},
\end{align}
correlates with a small relative change between successive outputs $\gamma_t:=\gamma_t^M$.
This insight underpins timestep-level caching strategies that leverage temporal smoothness to accelerate diffusion inference, a concept that is directly linked to our findings.

%% file: src/4-Method.tex
\section{Method}

\subsection{\our}
Our initial approach aimed to optimize the runtime performance of self-attention by incorporating the sparse online softmax technique from SpargeAttention.
As discussed in Section~\ref{ss:perm_attn}, whenever the condition~\eqref{eq:sparge_cond} is satisfied, the corresponding tile can be safely skipped.
This algorithm, which we refer to as \textit{PV-Skip}, terminates the tile-processing iteration early once this condition is met.
In practice, this requires computing the $QK$ product and its row-wise maximum, but allows us to omit both the element-wise exponentiation and the subsequent $PV$ product, as illustrated in Algorithm~\ref{alg:pv_skip}.\footnote{The epilogue of the online softmax is omitted since it is not pertinent to our discussion.}
Overall, this approach can reduce the computational cost of a skipped iteration by roughly half, provided that the skipping mechanism is effectively leveraged.
\begin{algorithm}
\caption{SpargeAttention \textit{PV-Skip}~\cite{zhang2025spargeattn}}\label{alg:cap}
\label{alg:pv_skip}
\begin{algorithmic}
\Require $Q_i,~\{K_j\},~\{V_j\}$
\While{$j$}
    \State $S_{ij} \gets Q_iK_j^\top / \sqrt{d}$
    \State $\mathbf{m}_\text{local} \gets \mathrm{rowmax}(S_{ij})$
    \State $\mathbf{m}_{ij} \gets \mathrm{rowmax}(\mathbf{m}_{i,j-1},\mathbf{m}_\text{local})$
    \If{$\max(\mathbf{m}_{\text{local}} - \mathbf{m}_{ij}) \le -\varepsilon$}
        \State \textbf{continue}
    \EndIf
    \State $\widetilde{P}_{ij} \gets \exp(S_{ij} - \mathbf{m}_{ij})$
    \State $\widetilde{O}_{ij} \gets \widetilde{P}_{ij} V_j$
    \State ...
\EndWhile
\end{algorithmic}
\end{algorithm}

We later observed that when the skipping condition was satisfied at timestep $t$, it tended to remain valid in subsequent timesteps as well.
Further investigation revealed that this temporal consistency persisted across timesteps for transformer blocks within the same layer.
Within each transformer block, computations are further partitioned by attention head.
Interestingly, we also found that the skipped tiles exhibited strong correlations across conditioning batches, suggesting that skip patterns could be inferred across batches at the same timestep.
A similar observation was reported by~\cite{lv2025fastercachetrainingfreevideodiffusion}.

\begin{figure}[h]
    \centering
    \includegraphics[width=.99\columnwidth]{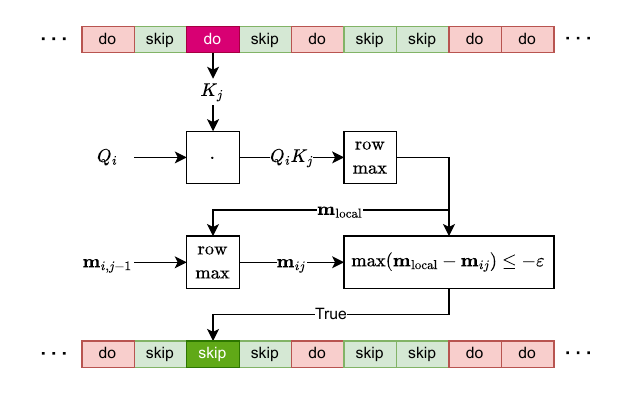} 
    % \vspace{-1em}
    \caption{A graphical depiction of the \textit{Skip-Mask} update step in Algorithm \ref{alg:qk_skip}.}
    \label{fig:skip_mask_update} 
\end{figure}

These observations led to the development of \our's \textit{QK-Skip} algorithm.
As shown in both Algorithm~\ref{alg:qk_skip} and Figure \ref{fig:skip_mask_update}, the method maintains a \emph{Skip-Mask} that is updated at each timestep.
As the diffusion process progresses, the number of tiles marked for skipping gradually increases.
The \textit{PV-Skip} mechanism can be optionally integrated into this algorithm; however, for a sufficient number of timesteps, its additional benefit becomes marginal.
\begin{algorithm}
\caption{\our \textit{QK-Skip}}\label{alg:cap}
\label{alg:qk_skip}
\begin{algorithmic}
\Require $Q_i,~\{K_j\},~\{V_j\},~SkipMask$
\While{$j$}
    \If{$SkipMask(i,j)$}
        \State \textbf{continue}
    \EndIf
    \State $S_{ij} \gets Q_iK_j^\top / \sqrt{d}$
    \State $\mathbf{m}_\text{local} \gets \mathrm{rowmax}(S_{ij})$
    \State $\mathbf{m}_{ij} \gets \mathrm{rowmax}(\mathbf{m}_{i,j-1},\mathbf{m}_\text{local})$
    \If{$\max(\mathbf{m}_{\text{local}} - \mathbf{m}_{ij}) \le -\varepsilon$}
        \State $SkipMask(i,j) \gets$ True
        \State \textbf{continue}
    \EndIf
    \State $\widetilde{P}_{ij} \gets \exp(S_{ij} - \mathbf{m}_{ij})$
    \State $\widetilde{O}_{ij} \gets \widetilde{P}_{ij} V_j$
    \State ...
\EndWhile
\end{algorithmic}
\end{algorithm}

\begin{figure}[h]
    \centering
    \includegraphics[width=.99\columnwidth]{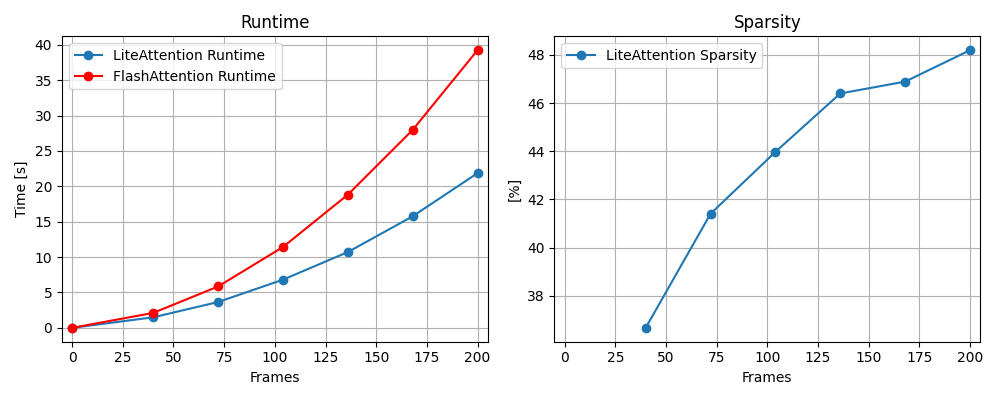} 
    % \vspace{-1em}
    \caption{Toy run of FlashAttention vs. \our within a video diffusion model for a varying number of video frames. \textbf{Left:} provides the runtimes. \textbf{Right:} provides \our's sparsity. If we assume that FlashAttention is of quadratic complexity, then this suggests that \our is of lower complexity, otherwise we would expect the sparsity percentage to be constant and not increasing.}
    \label{fig:lite_attn_perf} 
\end{figure}

Finally, we empirically observe that \our exhibits sub-quadratic complexity.
This claim is supported by a toy experiment, with results presented in Figure~\ref{fig:lite_attn_perf}.
We evaluated both FlashAttention and \our within a video diffusion model across a varying number of video frames.
Assuming FlashAttention scales quadratically, the observed trend suggests that \our operates with lower effective complexity - otherwise, the skip percentage would remain approximately constant rather than increasing with sequence length.

\subsection{From Caching to Skipping (Informal)}
Building on prior work on caching schemes, we can draw an informal connection between those observations and our findings.
Specifically, we argue that the slow evolution of transformer outputs across timesteps is closely related to the gradual evolution of the transformer's transition matrix $P$.

Our analysis is qualitative and omits certain architectural details.
In particular, we consider a single-headed transformer whose output is the self-attention output, ignoring contributions from the MLP layer.
Let us denote the transformer output as
\begin{align}
    \mathbf{y}_t = \mathbf{p}_t V_t,
\end{align}
where $\mathbf{p}_t$ is a row of the transition matrix $P$ and $\mathbf{y}_t$ is an output token.
We will argue that small changes in $\mathbf{p}$ imply small changes in $\mathbf{y}$ and vise-versa.

A direct computation of the output difference gives
\begin{align}
    \Delta \mathbf{y} 
    &= \mathbf{y}_t - \mathbf{y}_{t-\delta} \\
    &= \mathbf{p}_t V_t - \mathbf{p}_{t-\delta} V_{t-\delta} \\
    &= (\mathbf{p}_t - \mathbf{p}_{t-\delta}) V_t + \mathbf{p}_{t-\delta} (V_t - V_{t-\delta}) \\
    &= \Delta \mathbf{p} V_t + \mathbf{p}_{t-\delta} \Delta V.
\end{align}

An upper bound on the output difference follows
\begin{align}
    \|\Delta \mathbf{y}\| \leq \|\Delta \mathbf{p}\| \|V_t\| + \|\Delta V\|,
    \label{eq:y_from_p}
\end{align}
where all norms are Euclidean (Frobenius for matrices).
This inequality follows from the triangle and Cauchy-Schwarz inequalities and the fact that $\|\mathbf{p}_{t-\delta}\| \leq 1$ since its entries are non-negative and sum to 1.

Conversely, the transition difference satisfies
\begin{align}
    \Delta \mathbf{p} = (\Delta \mathbf{y} - \mathbf{p}_{t-\delta} \Delta V) V_t^\dagger,
\end{align}
where $V_t^\dagger$ is the Moore-Penrose pseudoinverse of $V_t$.
The transition difference can be bounded as
\begin{align}
    \|\Delta \mathbf{p}\| \leq \frac{\|\Delta \mathbf{y}\| + \|\Delta V\|}{\sigma_{\min}(V_t)},
    \label{eq:p_from_y}
\end{align}
where $\sigma_{\min}(V_t)$ is the smallest singular value of $V_t$.

The forward relation \eqref{eq:y_from_p} suggests that small perturbations in the transition matrix induce small perturbations in the transformer's output.
The reverse relation is less clear unless $V_t$ is well-conditioned, i.e., its smallest singular value is not too small.
Nevertheless, our skipping scheme only relies on the slow-evolution assumption to bind the tiles eligible for skipping.
Tiles that are not skipped are recomputed from scratch and may undergo large changes.

Finally, we note that an additional advantage of transition-based skipping over transformer output caching is the significantly reduced intermediate memory requirements.

\subsection{Skipping Condition}
\label{sec:skip_cond}
The skipping condition~\eqref{eq:sparge_cond} is \emph{local} in nature, as it evaluates each tile's individual contribution to the output while disregarding interactions across multiple tiles.
In contrast, a \emph{global} condition would typically assess the entire row of tiles and omit those whose cumulative contribution falls below a specified threshold (for example, removing the weakest tiles such that their cumulative sum does not exceed a given bound).
Naturally, the local condition is somewhat more conservative than the global one, as it must account for cases where several tiles might be simultaneously omitted, even though in practice only a few are.
Nevertheless, our experiments indicate that the proposed local criterion, when properly calibrated, performs similarly to several global conditions we evaluated.

In addition to the locality of the condition~\eqref{eq:sparge_cond}, its precision is further limited by the use of the cumulative row maximum $\mathbf{m}_{ij}$ instead of the true global row maximum.
This limitation arises from the causal nature of the online softmax algorithm.
Motivated by \cite{li2025radial}, we explored alternative orderings of the $j$-loop within the attention kernel.
We observed that radial-centric ordering improved performance by reaching the global maximum faster.

% \begin{figure}[h]
%     \centering
%     \includegraphics[width=.81\columnwidth]{src/figs/timestep_error.pdf} 
%     % \vspace{-1em}
%     \caption{Analysis of the final attention errors using a fixed error for different timesteps.}
%     \label{fig:timestep_error} 
% \end{figure}

\subsection{Accumulated-Error Calibration}
Diffusion models usually require multiple timesteps~\citep{wan2025wan} to progressively denoise an entire data sample. We note that attention errors brought about by sparsity at different timesteps have varying impacts on the final attention output (i.e., at the last timestep): the earlier the timestep, the greater its influence on the final output error. As shown in Table~\ref{tab:timestep_error}, we analyze this effect on the Wan2.1~\citep{wan2025wan} model and find that, under the same attention error magnitude, earlier timesteps contribute more significantly to the final attention error. Based on this observation, we propose assigning different error bounds to different timesteps and searching for the optimal \textit{PV-threshold} for each timestep. 

Specifically, we divide the timesteps evenly into three segments. 
For these three segments, we set the error bounds to $\xi - \tau$, $\xi$, and $\xi + \tau$, respectively. 
Based on these error bounds, we search for the optimal \textit{PV-threshold} per each timestep. 
Here, the error is defined as the relative L1 error $\eta_t= {|O_t^s - O_t|}/{|O_t|}$, where $|\cdot|$ denotes the matrix $L_1$ norm, $O_t^s$ represents the output of the sparse attention, and $O_t$ represents the output of full attention, all at timestep $t$.

\begin{table}[h!]
\centering
\caption{Analysis of the final attention error $\eta_{49}$ for a fixed \textit{PV-threshold} at different intermediate timesteps.}
\begin{tabular}{ccccc}
\toprule
\textbf{Timestep} & \textbf{0} & \textbf{16} & \textbf{32} & \textbf{48} \\
\midrule
$\eta_{49}$ & 0.392 & 0.375 & 0.325 & 0.318 \\
\bottomrule
\end{tabular}
\label{tab:timestep_error}
\end{table}

%% file: src/6-Experiments.tex
\section{Experiments}

\subsection{Setup}
\smallskip \noindent{\textbf{Models, Dataset, and Baselines.} 
We evaluate \our using the 12 prompts dataset from OpenSora1.0~\cite{zheng2024opensora}. 
For video generation models, we consider Wan2.1-14B and Wan2.2-14B~\citep{wan2025wan}. 
FlashAttention3 (FA3)~\citep{shah2024flashattention}, SparseVideoGen (SVG)~\cite{xi2025svg}, and RadialAttention (Radial)~\cite{li2025radial} comparison baseline.

\smallskip
\noindent{\textbf{Metrics.} 
Generated video quality is evaluated using VBench~\cite{huang2024vbench} across the metrics \textit{Aesthetic Quality} (AQ), \textit{Background Consistency} (BC), \textit{Dynamic Degree} (DD), \textit{Imaging Quality} (IQ), \textit{Subject Consistency} (SC), \textit{Temporal Flickering} (TF), and \textit{Temporal Style} (TS). 
All values are averaged over the dataset.
For \our, \textit{sparsity} (Sps) denotes the fraction of computations skipped relative to full attention, averaged over the generation process. 
For all other methods, we report the sparsity values they report.

% \begin{table}[h]
% \centering
% \caption{Summary of VBench evaluation metrics.}
% \begin{tabular}{lll}
% \toprule
% \textbf{Metric} & \textbf{Abb.} & \textbf{Description} \\
% \midrule
% \makecell[l]{Aesthetic \\ Quality} & AQ & \makecell[l]{Visual appeal of frames:\\ composition, color, and style.} \\
% \makecell[l]{Background \\ Consistency} & BC & \makecell[l]{Stability and coherence of\\ the background across frames.} \\
% \makecell[l]{Dynamic \\ Degree} & DD & \makecell[l]{Level of motion or activity\\ in the video.} \\
% \makecell[l]{Imaging \\ Quality} & IQ & \makecell[l]{Technical quality: sharpness,\\ clarity, and artifact-free.} \\
% \makecell[l]{Subject \\ Consistency} & SC & \makecell[l]{Main subjects remain visually\\ consistent across frames.} \\
% \makecell[l]{Temporal \\ Flickering} & TF & \makecell[l]{Absence of abrupt changes or\\ flickers between frames.} \\
% \makecell[l]{Temporal \\ Style} & TS & \makecell[l]{Consistency of visual or motion\\ style throughout the video.} \\
% \bottomrule
% \end{tabular}
% \label{tab:vbench2_metrics}
% \end{table}

\smallskip \noindent{\textbf{Settings.}
Throughout our experiments, \our was applied exclusively to accelerate the self-attention primitive and was run using the standard (suboptimal) linear ordering for the $j$-loop (see Section~\ref{sec:skip_cond}).

The results in Table~\ref{tab:compare_results} were obtained using calibrated \textit{PV-thresholds} with $\tau = 0.01$ and $\xi=0.075$.
%$\xi=0.065$ for LTX-13B 
% For Wan2.2-14B, we observed that the \textit{Skip-List} resets after $25$ timesteps (out of $40$) due to a change in the self-attention dimensions induced by the model's behaviour.
% While this issue could potentially be mitigated through adjustment, we did not attempt to correct it in our experiments. 

The results in Table~\ref{tab:ablation_results} were generated for Wan2.1 without calibration. 
Instead, the \textit{PV-threshold} was set to $-8$ for the first $20$ timesteps, and a grid search was used to select thresholds for the last $30$ timesteps to achieve the desired sparsity level. Experiments were conducted on NVIDIA's H200 GPU.

\subsection{Results}
\smallskip
\noindent{\textbf{Effectiveness.}
Table~\ref{tab:ablation_results} examines the impact of increasing sparsity on video quality. 
At $77\%$ sparsity, we observed visible distortion in the generated video, which is empirically reflected in the TS metric. 
Comparing this metric in Table~\ref{tab:compare_results} indicates that \our preserves visual quality comparable to full attention (FA3), while SVG and Radial show marked degradation.

\smallskip
\noindent{\textbf{Efficiency.}
Although SVG and Radial report higher nominal sparsity in Table~\ref{tab:compare_results}, \our achieves at least $10\%$ greater runtime improvement. 
Combined with their stronger quality drop, this highlights \our's superior trade-off between efficiency and fidelity. 
We expect an additional $10$–$20\%$ gain in sparsity without quality loss once optimized $j$-loop ordering is applied.
Notably, Table~\ref{tab:ablation_results} shows that runtime reduction scales nearly one-to-one with skipped computation.

\smallskip
\noindent{\textbf{Ablation Study.}}
Table~\ref{tab:ablation_results} further examines quality and runtime under varying sparsity levels using an uncalibrated setup, where the \textit{PV-threshold} was increased over the last 30 of 50 timesteps. 
Video quality degrades sharply beyond $70\%$ sparsity. 
Comparing the $42\%$ entry in Table~\ref{tab:ablation_results} with the corresponding entry in Table~\ref{tab:compare_results} for Wan2.1-14B shows that calibrated runs achieve substantially higher quality at the same sparsity. 
This suggests that with proper calibration and improved $j$-loop ordering, sparsity, and thus runtime, can reach around $70\%$ without visible quality loss.

%% file: src/7-Conclusions.tex
\section{Conclusions}
We presented \our, a method that exploits the temporal coherence of sparsity in diffusion transformer attention to accelerate video generation. 
By identifying and propagating skippable tiles across timesteps, \our combines the adaptivity of dynamic sparsity with the efficiency of static approaches, achieving substantial runtime reductions without compromising video fidelity. 
The method integrates seamlessly into existing CUDA-accelerated attention kernels, requires no model retraining, and demonstrates that evolutionary computation skips can unlock practical, high-performance DiT inference at scale.

\newpage